\documentclass[letterpaper, 10 pt, conference]{IEEEtran}

\IEEEoverridecommandlockouts
\usepackage{cite}
\usepackage{amsmath,amssymb,amsfonts}
\usepackage{algorithmic}
\usepackage{graphicx}
\usepackage{textcomp}
\usepackage{xcolor}
\usepackage{booktabs}
\usepackage{stfloats}
\usepackage{multirow}
\usepackage{makecell}
\usepackage{url}
\usepackage{hyperref}

\def\BibTeX{{\rm B\kern-.05em{\sc i\kern-.025em b}\kern-.08em
    T\kern-.1667em\lower.7ex\hbox{E}\kern-.125emX}}
\begin{document}

\title{Fast Policy Learning for 6-DOF Position Control of Underwater Vehicles}

\author{\IEEEauthorblockN{1\textsuperscript{st} Sümer Tunçay}
\IEEEauthorblockA{\textit{School of Engineering and Physical Sciences} \\
\textit{Heriot-Watt University}\\
Edinburgh, United Kingdom \\
st2126@hw.ac.uk}
\and
\IEEEauthorblockN{2\textsuperscript{nd} Alain Andres}
\IEEEauthorblockA{\textit{TECNALIA, BRTA} \\
San Sebastian, Spain \\
alain.andres@tecnalia.com}
\and
\IEEEauthorblockN{3\textsuperscript{rd} Ignacio Carlucho}
\IEEEauthorblockA{\textit{School of Engineering and Physical Sciences} \\
\textit{Heriot-Watt University}\\
Edinburgh, United Kingdom \\
ignacio.carlucho@hw.ac.uk}
}
\maketitle

\begin{abstract}

Autonomous Underwater Vehicles (AUVs) require reliable six–degree-of-freedom (6-DOF) position control to operate effectively in complex and dynamic marine environments. Traditional controllers are effective under nominal conditions but exhibit degraded performance when faced with unmodeled dynamics or environmental disturbances. Reinforcement learning (RL) provides a powerful alternative but training is typically slow and sim-to-real transfer remains challenging. This work introduces a GPU-accelerated RL training pipeline built in JAX and MuJoCo-XLA (MJX). By jointly JIT-compiling large-scale parallel physics simulation and learning updates, we achieve training times of under two minutes.
Through systematic evaluation of multiple RL algorithms, we show robust 6-DOF trajectory tracking and effective disturbance rejection in real underwater experiments, with policies transferred zero-shot from simulation.

\end{abstract}


\section{Introduction}

Autonomous Underwater Vehicles (AUVs) play a central role in modern marine robotics, supporting a wide range of marine applications, including the inspection and maintenance of offshore infrastructures and subsea pipeline and cables. Successful deployment of AUVs in complex missions depends on reliable 6-DOF position control. However, AUVs are subject to nonlinear and coupled hydrodynamic forces, which are often difficult to model with precision, making the task difficult. In addition, they operate in uncertain and highly dynamic environments, where disturbances such as ocean currents, turbulence, and wave-induced motions can significantly perturb the vehicle’s trajectory.

Conventional control approaches, such as Proportional-Integral-Derivative (PID) \cite{MIC-1996-1-5} and Model Predictive Control (MPC) \cite{ZHANG2019106309} have long been applied to underwater vehicles. PID controllers are simple and robust under nominal conditions but require frequent manual fine-tuning in the presence of strong coupling or unmodeled dynamics. MPC explicitly optimizes control inputs under system constraints, but its effectiveness depends on model fidelity, and even small discrepancies or unmodeled external disturbances can severely degrade performance. These limitations motivate the search for adaptive alternatives.

Reinforcement learning (RL) offers such an alternative by directly optimizing control policies through interaction with the environment. By doing so, and by leveraging the representation power of neural networks, RL policies can learn to handle non-linear and coupled dynamics. In practice, however, interactions are largely carried out in simulation, since real interactions are expensive, time-consuming and potentially hazardous. This reliance on simulation introduces the challenge of sim-to-real transfer, where policies that perform well in simulation may fail when deployed on real vehicles. Domain randomization \cite{Tobin2017DomainRF} and related techniques \cite{ANDRES2025112090} mitigate this gap by exposing policies to variability during training. Nevertheless, RL remains data-hungry and training times are often prohibitively long. Consequently, prior work on 6-DOF control of underwater vehicles has struggled to achieve strong performance under practical training times, limiting its deployment  on real platforms.

\begin{figure}[t]
    \centering
    \includegraphics[width=\columnwidth]{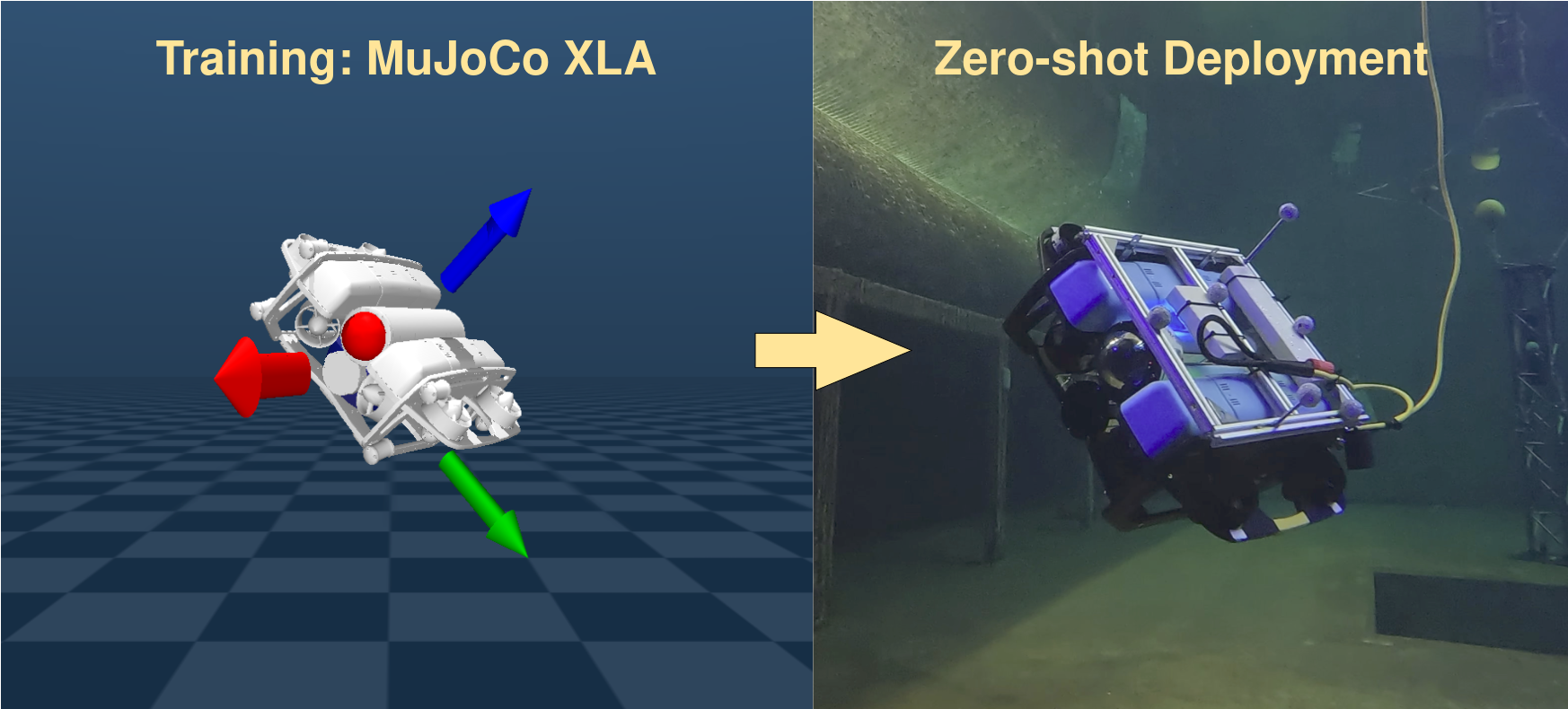}
    \caption{Execution of trained policies in simulation (MuJoCo visualizer, left) and on the real vehicle (right).}
    \label{fig:bluerov_images}
\end{figure}

In this work, we introduce a GPU-accelerated training pipeline for the 6-DOF underwater vehicle position control task built in JAX and MuJoCo-XLA (MJX), a differentiable variant of MuJoCo \cite{6386109} implemented in JAX. Our framework leverages two complementary acceleration mechanisms. First, massively parallelized simulation is achieved by vectorizing thousands of rollouts and executing them concurrently on the GPU. Second, just-in-time (JIT) compilation in JAX eliminates Python overhead by fusing physics simulation and learning updates into XLA-compiled functions \cite{50530}. We demonstrate that this parallelization, together with a concise state representation and basic domain randomisation, is enough to achieve zero-shot transfer for RL policies.

Our main contributions are:
\begin{itemize}
    \item Development of a fully JAX and MJX-based training pipeline for underwater vehicle control, supporting GPU-parallelized differentiable simulation which enables obtaining accurate control policies within few minutes of training.
    \item Evaluation of three state-of-the-art RL algorithms representing different paradigms: on-policy Proximal Policy Optimization (PPO) \cite{Schulman2017PPO}, off-policy Soft Actor-Critic with enhanced regularization (SAC-DroQ)\cite{hiraoka2022droq}\cite{Haarnoja2018SoftAO}, and differentiable-simulation–based Short Horizon Actor-Critic (SHAC) \cite{xu2021shac}, for the 6-DOF position control task. This includes a comparative analysis of their learning dynamics, control performance, and scaling behavior under parallelization.
    \item Experimental validation of trained policies on a real AUV in challenging tasks requiring precise 6-DOF tracking and resilience to external disturbances, demonstrating robust sim-to-real transfer. Performance is further compared against an MPC baseline under the same experiment setups.
\end{itemize}

Overall, our results highlight the effectiveness of the modern JAX/MJX stack as a scalable and efficient platform for fast policy learning with robust sim-to-real transfer. The proposed framework reduces policy training times to just a few minutes through combining large-scale GPU-parallelized differentiable simulation with JIT compilation. The trained policies transfer robustly to a real vehicle, outperforming an MPC baseline for 6-DOF position control task. 

\section{Related Work}

Several studies have explored RL for AUV control, but real-world demonstrations remain scarce and usually restricted to partial tasks rather than full 6-DOF control. In simulation, early work applied DDPG \cite{Lillicrap2015ContinuousCW} to simplified problem settings. In \cite{8604791}, the authors demonstrated 3D translational position control, while in \cite{FANG2022110452}, the same approach was extended to yaw regulation as well. In \cite{s22166072}, the authors proposed a SAC-based controller for 6-DOF regulation, but the evaluation was confined to simulation and did not demonstrate complex maneuvers requiring accurate orientation control. A more comprehensive study was presented in \cite{10.1007/978-3-032-01486-3_29}, where several RL algorithms were compared against an MPC baseline on the full 6-DOF position control task, although this work also remained limited to simulation.

Several frameworks have been proposed to accelerate training. In \cite{10682328}, a Gazebo-based system with multithreading reduced training times to around 40 minutes, but no real experiments were shown. Authors of  \cite{chu_2025_MarineGymHighPerformanceReinforcement} introduced an IsaacLab plugin for hydrodynamics, enabling large-scale parallelization, but again without sim-to-real validation.

A smaller set of works demonstrated real experiments. In \cite{CARLUCHO201871}, DDPG was applied for 6-DOF velocity control and successfully deployed on a physical vehicle. In \cite{WENG2025121047}, SAC was used for trajectory tracking, but the evaluation was restricted to translational degrees of freedom. In \cite{app13031723}, data-informed domain randomization enabled sim-to-real transfer for translational position and yaw control. Some studies have focused on specific objectives requiring position control. For example, \cite{blabla} applied SAC to a 3DoF link-alignment task, demonstrating moderate performance on a real vehicle. In \cite{Cai2024LearningTS}, authors demonstrate real experiments on a station keeping task, where PPO policies are trained in IsaacLab within 15 minutes by leveraging parallelized GPU-based simulation. However, station keeping task is not fully representative of 6-DOF control capabilities. In \cite{11020757}, the authors present an end-to-end DRL controller for trajectory following with full 6-DOF control, trained over 15 hours and validated in both simulation and in-water experiments. 

In summary, RL has shown recent success in 6-DOF AUV control, however prior work mostly remains confined to simulation, and in-water validated approaches are either trained over impractical timescales or shown limited control performances.

\section{Background}
Reinforcement learning models decision-making as an agent interacting with an environment via its policy. At each step, the agent selects an action $a_t$ from the current state $s_t$, receives a reward $r_t$, and transitions to a new state $s_{t+1}$. This process is formalized as a Markov Decision Process (MDP) by a 5-tuple $\{\mathcal{S}, \mathcal{A}, \mathcal{T}, \mathcal{R}, \gamma\}$, where $\mathcal{S}$ stands for the state space, $\mathcal{A}$ the action space, $\mathcal{T}$ the transition probability model, $\mathcal{R}$ the reward function, and $\gamma \in [0,1]$ the discount factor. The objective is to find a policy $\pi$ that maximizes the expected discounted return $J(\pi) = \mathbb{E}_{\pi}\left[\sum_{t=0}^\infty \gamma^t r_t \right]$.

\subsection{Algorithms Considered for Continuous Control}
In the following, we briefly introduce three reinforcement learning algorithms for continuous control from different paradigms; on-policy, off-policy and differentiable-simulation-based learning. 

\subsubsection{On-Policy (PPO)}
On-policy methods update the policy only with fresh trajectories generated from the current policy, which typically results in lower sample efficiency but more stable learning dynamics. PPO \cite{Schulman2017PPO} is one of the most widely used algorithms in this family due to its simplicity and robustness. It introduces a clipped surrogate objective that prevents destructive policy updates by limiting the change in the probability ratio between old and new policies.

\subsubsection{Off-Policy (SAC-DroQ)}
Off-policy methods optimize a policy using replayed experience rather than only fresh trajectories, which makes them significantly more sample efficient than on-policy approaches. A prominent representative approach for continuous problems is SAC \cite{haarnoja2018soft}, which maximizes a maximum-entropy objective that encourages both reward and policy stochasticity. In this study, we consider DroQ\cite{hiraoka2022droq}, which builds upon SAC by employing a small ensemble of Q-networks that are regularized with dropout and layer normalization. This combination enables stable learning at higher update-to-data ratios by mitigating overestimation bias and reducing overfitting to the target networks. As a result, DroQ demonstrates high sample efficiency.

\subsubsection{Differentiable Simulation Based (SHAC)}
A third class of approaches utilizes differentiable simulation to leverage first order gradients by backpropagating gradients through the system dynamics to improve sample efficiency. Backpropagating over the whole trajectory results in unstable learning due to exploding and vanishing gradients, so short sub-trajectories are used in practice instead. SHAC \cite{xu2021shac} builds upon this idea and introduces a value network that provides terminal value estimates at the end of short rollout windows, which improve the policy update by combining short-horizon rewards within the window and the critic’s terminal estimate. 

Together, these methods span distinct paradigms in continuous control. PPO as a stable and widely adopted on-policy approach, DroQ as a state-of-the-art off-policy approach with high sample efficiency, and SHAC as a differentiable-simulation method that exploits gradients through dynamics. This selection allows us to compare learning performance across different algorithmic families.

\section{Method}

\subsection{GPU-based Parallelized Simulation}

Although significant progress on data efficiency has been made in the literature, requirement of large amounts of interaction data remains to be the bottleneck to train RL policies. To make training feasible within a reasonable wall-clock time, we leveraged MJX, which provides a JAX-compatible interface for rigid body simulation. 

Using MJX, we implemented a parallelized training environment where multiple instances of the AUV simulator are executed in batch on a single GPU. This allows thousands of rollouts to be simulated simultaneously with minimal overhead by fully exploiting modern GPU vectorization. Each environment evolves independently, with its own randomized initial conditions and domain randomization parameters, but shares the same physics model and integrator. On an \textit{NVIDIA GeForce RTX 4060}, our setup scaled training up to 4096 parallel environments. Additionally, by applying jit compilation to both the environment rollouts and the learning updates, we achieve substantial acceleration in training, enabling high-performing policies to be obtained within \textit{minutes}. 

\subsection{Environment Specifications}

The training objective is 6-DOF position control: given a target pose, the agent must regulate translational and rotational errors simultaneously, producing corrective actions that stabilize the vehicle at the commanded setpoint. The environment models a free–floating AUV with full 6-DOF dynamics. We set up the system in MJX, where the vehicle is represented as a single free body with ellipsoid geometry, actuated by virtual motors that directly apply forces and torques on its free joint. For hydrodynamic effects, we used the inertia fluid drag model provided by MJX, with fluid density and viscosity parameters set to match water. In this model, each body is represented by an equivalent inertia box derived from its mass and inertia tensor. Hydrodynamic forces and torques are computed as quadratic drag and viscous resistance on the equivalent inertia box. Buoyancy force is not modeled explicitly, rather it is approximated by modifying the effective gravity, assuming centers of gravity and buoyancy are aligned.

\subsubsection{Observation Space}

The choice of observation space is critical for both learning performance and stability. Many formulations are possible: one can expose the policy to absolute system states together with the commanded references, or even augmented with the previous actions taken. However, for the problem of 6-DOF position control, such formulations increase the input dimensionality and place the burden of computing control errors on the neural network. Instead, for positions, we adopt an \emph{error-based observation space}, where the policy directly receives the difference between the current state and the desired reference. We also extend the observations with body velocities. Specifically, the observation vector is composed of:
\begin{itemize}
\item Position error expressed in the body frame ($\Delta x, \Delta y, \Delta z$),
\item Attitude error represented as a 3D axis--angle vector between the reference and the measured orientation,
\item Linear velocity in the body frame ($u, v, w$),
\item Angular velocity in the body frame ($p, q, r$).
\end{itemize}

This results in a 12-dimensional observation vector. An additional consideration is whether to normalize observations. While normalization is often employed in RL to avoid disproportionate influence of inputs with larger numerical ranges, in our case the error-based formulation naturally yields inputs of comparable scale, and thus we chose not to apply normalization.

\subsubsection{Action Space}

The policy outputs a continuous 6-dimensional action vector:

\begin{equation}
\mathcal{A} = \{F_x, F_y, F_z, \tau_\phi, \tau_\theta, \tau_\psi\},
\end{equation}

\noindent corresponding to commanded forces in surge ($F_x$), sway ($F_y$), and heave ($F_z$), and torques around the body-fixed roll ($\tau_\phi$), pitch ($\tau_\theta$), and yaw ($\tau_\psi$). Each action component is normalized to the range $[-1,1]$. At runtime, normalized actions are scaled to maximum allowable values $F_{max}$ and $\tau_{max}$. Agent operates in terms of body-fixed forces and torques rather than individual thruster commands. This abstraction reduces the sim-to-real gap, since the learned policy is not tied to a specific thruster configuration or motor dynamics. The burden is deferred to deployment, where these actuator constraints must be reintroduced through a thruster allocation model and motor dynamics, but this separation makes training faster and more general while preserving the ability to adapt to different hardware. 

It is important to notice that both observations and actions are expressed in the body-fixed frame. This alignment simplifies the policy’s task, since the agent does not need to learn frame transformations between the world and body coordinates. Instead, it directly associates body-frame errors with the corrective forces and torques that act in the same frame. This reduces representational complexity and improves training stability.

\subsubsection{Reward Function}
The reward function ($\mathcal{R}$) is designed to encourage accurate 6-DOF control while discouraging aggressive control actions. At each timestep, the agent receives a scalar reward $r_t$:
\begin{equation}
    r_t =  r_{\text{pos}} + r_{\text{att}} + r_{\text{act}} + r_{\text{vel}} + r_{\text{act-mavg}},
\end{equation}
where each term aim on addressing a different aspect of the agent's behavior. Position tracking is enforced by penalizing the squared norm of the body-frame position error, weighted anisotropically:
    \begin{equation}
        r_{\text{pos}} = - w_{\text{pos}} \, \big\lVert D \, \Delta \mathbf{p} \big\rVert^2,
    \end{equation}
with $\Delta \mathbf{p} = (\Delta x, \Delta y, \Delta z)$ and $D = \text{diag}(w_x, w_y, w_z)$. This allows different importance to be assigned along surge, sway, and heave. Orientation deviation is penalized via the squared norm of the axis-angle error vector:
    \begin{equation}
    r_{\text{att}} = - w_{\text{att}} \, \lVert \Delta \boldsymbol{\theta} \rVert^2.
    \end{equation}
To discourage large torques, we penalize the squared norm of the torque command vector:
    \begin{equation}
    r_{\text{act}} = - w_{\text{act}} \, \lVert \boldsymbol{\tau} \rVert^2,
    \end{equation}
where $\boldsymbol{\tau} = (\tau_\phi, \tau_\theta, \tau_\psi)$. Squared norm of angular velocities are penalized to discourage oscillations and aggressive rotations:
    \begin{equation}
    r_{\text{vel}} = - w_{\text{vel}} \, \lVert \boldsymbol{\omega} \rVert^2,
    \end{equation}
with $\boldsymbol{\omega} = (p,q,r)$. Finally, to promote smoother controls, we penalize deviations of the current action from the moving average of past actions. 
    \begin{equation}
        r_{\text{act-mavg}} = - w_{\text{act-mavg}} \, \lVert \mathbf{a}_t - \bar{\mathbf{a}} \rVert^2,
    \end{equation}
    where $\bar{\mathbf{a}}$ denotes the mean of the recent action history. This term is particularly important for sim-to-real transfer, as it discourages the agent from learning bang–bang strategies characterized by high-magnitude, high-frequency control inputs. Such behaviors may be tolerated in simulation but are detrimental on real hardware, where actuator dynamics and power limitations make smooth control essential.

\subsection{Domain Randomization}

In order to improve robustness and enable sim-to-real transfer, we applied a simple domain randomization where the gravitational acceleration was randomized for each environment to approximate variations in buoyancy, under the assumption that the centre of buoyancy and the centre of gravity are aligned. This procedure captures variability in effective buoyant forces without explicitly modelling off-centre buoyancy effects, which remain fixed in our setup. 

\subsection{Model Predictive Controller Baseline}

As a model-based baseline, we deployed an MPC scheme for 6-DOF position control \cite{10.1007/978-3-032-01486-3_29}. The problem is formulated as a constrained quadratic program over a finite prediction horizon, where a quadratic cost penalizes both state deviation and control effort:

\begin{equation}
\begin{aligned}
\min_{x,u} \quad J = \sum_{k=0}^{H-1} & \; \tfrac{1}{2}\big(x_k^{T} Q x_k + u_k^{T} R u_k\big) + x_H^{T} Q_H x_H \\
\text{s.t.} \quad & x_{k+1} = f(x_k,u_k), \quad k = 0,\dots,H-1, \\
& \underline{x} \leq x_k \leq \bar{x}, \quad k = 0,\dots,H, \\
& \underline{u} \leq u_k \leq \bar{u}, \quad k = 0,\dots,H_c-1, \\
& u_k = u_{H_c-1}, \quad k = H_c,\dots,H-1.
\end{aligned}
\end{equation}

Here, $Q$, $R$, and $Q_H$ denote state, control, and terminal cost matrices. The parameters $H$ and $H_c$ define the prediction and control horizons. State and control constraints are imposed by $\underline{x}, \bar{x}$ and $\underline{u}, \bar{u}$. 

The vehicle dynamics $f(x,u)$ follow the non-linear 6-DOF equations of motion described by Fossen \cite{fossen1994guidance}. To enable efficient linearization at each control step, the dynamics were implemented in PyTorch \cite{Paszke2019PyTorchAI}. This allowed us to exploit automatic differentiation: the Jacobians $\partial f / \partial x$ and $\partial f / \partial u$ are computed on the fly using the \textit{jacobian} operator. These linearized dynamics define a local approximation of the system around the current operating condition, yielding a quadratic program which we solve efficiently with CVXPY \cite{Diamond2016CVXPYAP}, outputting the commanded wrench.

\section{Experiments}

We conduct a comprehensive set of experiments to evaluate the effectiveness of the proposed reinforcement learning framework for 6-DOF position control of underwater vehicles. The evaluation is structured into simulation-based training experiments and real-world deployment trials. In simulation, we assess training efficiency, convergence behavior, and sensitivity to parallelization and attitude representation. The real-world evaluation focuses on assessing the sim-to-real transfer and control effectiveness of the learned policies on a physical AUV, including both trajectory tracking and external disturbance rejection scenarios.
All the RL algorithms have been implemented in JAX/Flax \cite{flax2020github}, which ensures that both the MJX physics simulation and the learning updates could be jointly compiled and executed on GPU. This yields a fully differentiable and end-to-end vectorized training pipeline. 

\subsection{Simulation Experiments}
\subsubsection{6-DOF RMSE Throughout Training} 
\begin{figure}[t]
    \centering
    \includegraphics[width=\columnwidth]{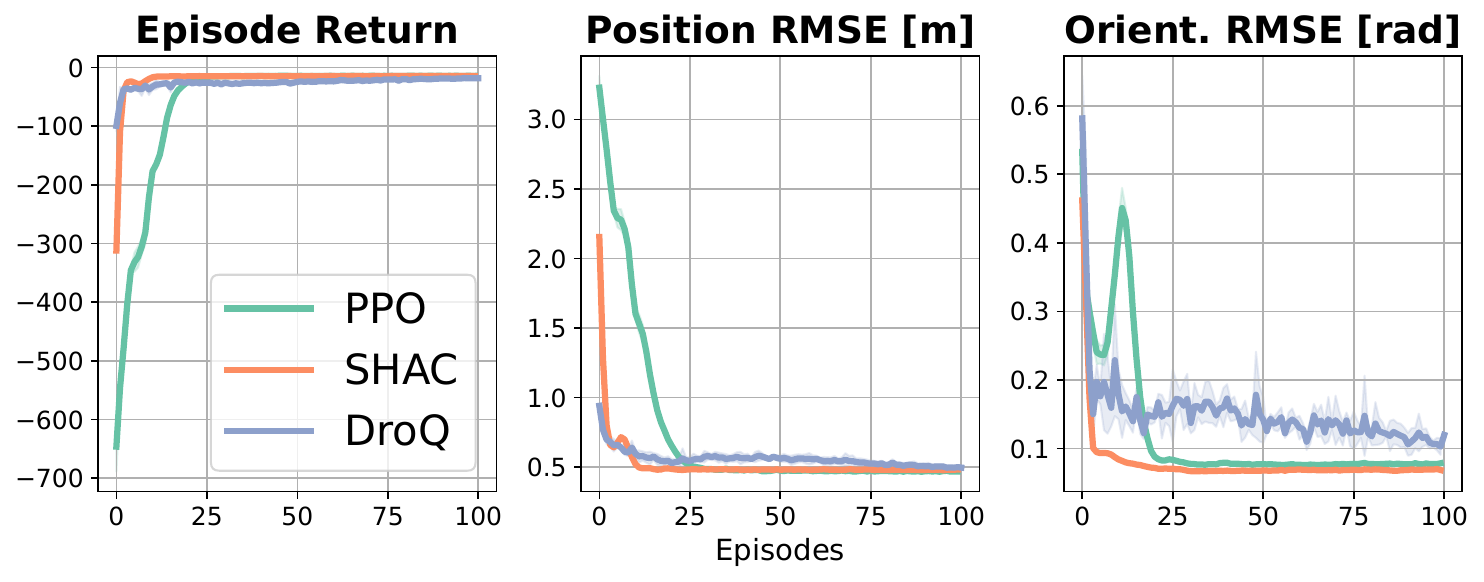}
    \caption{Returns (left), episodic position (middle) and orientation (right) RMSEs during training using 4096 environments for SHAC, PPO and DroQ}
    \label{fig:err_graphs}
\end{figure}

We begin by analyzing training performance using the maximum parallelization setting of 4096 environments, which represents the upper limit of our hardware. Each experiment was run for 100 episodes across three random seeds, with each episode consisting of 512 environment steps. At every episode, a random reference is sampled within $[-2,2]m$ and $[-60,60]^\circ$. Figure \ref{fig:err_graphs} summarizes the results, reporting episodic returns alongside position and orientation RMSE over the course of training. All three algorithms achieve stable 6-DOF control under this setup, with PPO and SHAC converging rapidly, while DroQ reduces orientation errors at a slower rate.

\subsubsection{Effect of Parallelized Environments}

\begin{figure}
    \centering
    \includegraphics[width=\columnwidth]{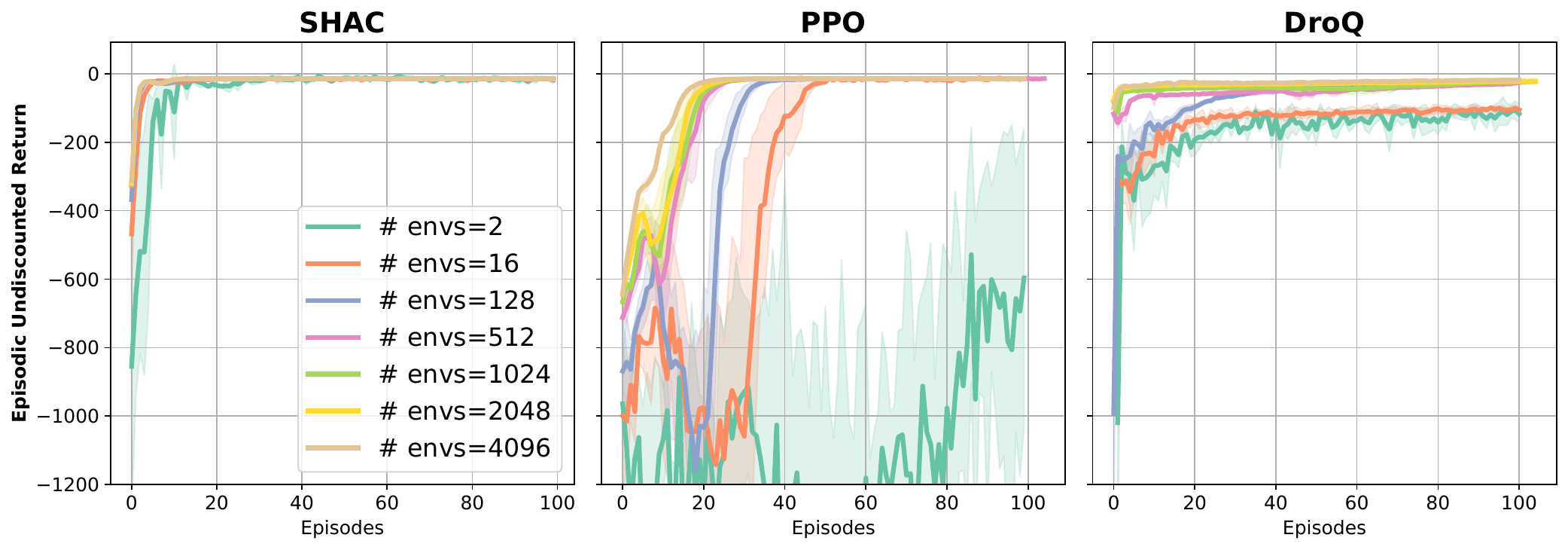}
    \caption{Effect of parallelized environments on episodic returns for SHAC (left), PPO (middle) and DroQ (right).}
    \label{fig:parallel_envs_performance}
\end{figure}

\begin{figure}
    \centering
    \includegraphics[width=\columnwidth]{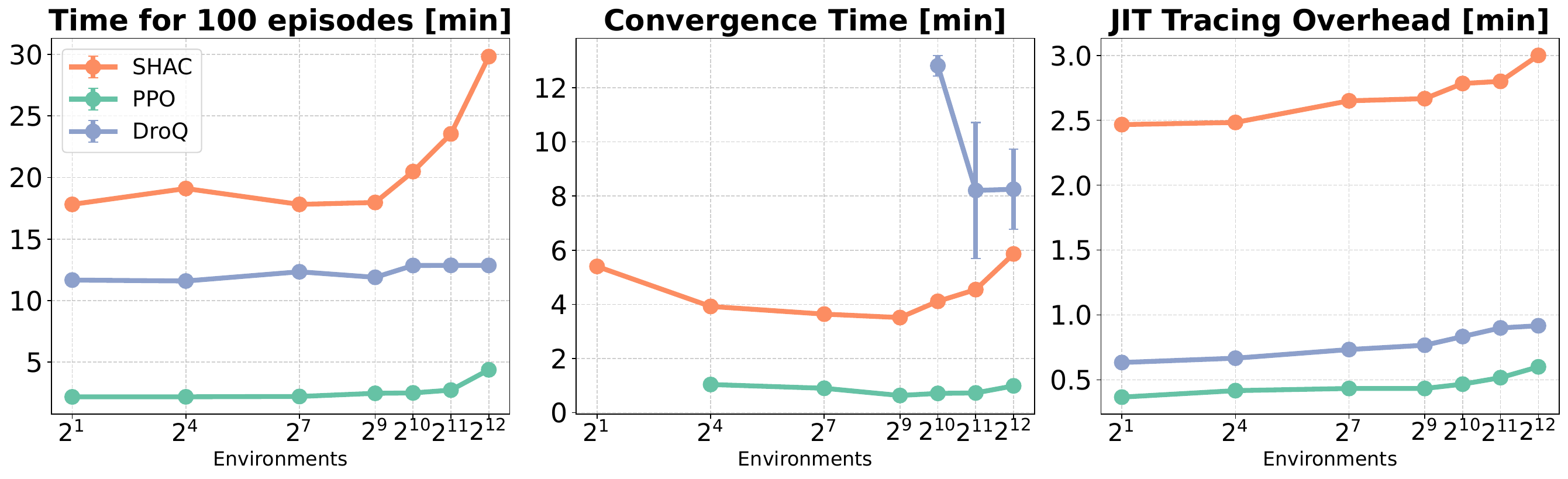}
    \caption{Effect of parallelized environments on total training time and convergence time. Presents time required to complete 100 episodes of training (left), convergence time (middle) and JIT tracing time (right).}
    \label{fig:parallel_envs_time}
\end{figure}

A central motivation of this work is to make policy training for underwater vehicle control feasible on practical time scales. To identify a time-optimal regime where wall-clock efficiency and convergence speed are jointly maximised, we performed training runs for each algorithm while varying the number of parallel environments. 

We report the mean wall-clock time required to complete 100 training episodes and one-time jit compilation overhead. Additionally, we report convergence speed, defined as the wall-clock time until the agent achieved stable 6-DOF position control, i.e., when tracking errors remained consistently bounded for the remainder of training. Our results,  averaged over three random seeds,  are presented in Figure \ref{fig:parallel_envs_performance} and Figure \ref{fig:parallel_envs_time}, which show how different reinforcement learning paradigms benefit from parallelisation in terms of both performance and total training time, respectively.

For the on-policy model-free PPO algorithm, scaling the number of environments proved highly effective: while convergence was not achieved within 100 episodes using only 2 environments, the agent consistently converged in under three minutes across all larger environment counts. In contrast, the differentiable model-based SHAC showed a more complex behavior. Firstly, due to its reliance on backpropagating through the physics model, training time for 100 episodes is significantly higher than the on-policy and off-policy counterparts. Nevertheless, the same reliance enables it to converge consistently within 3-6 minutes across all environments. A sharp increase in total training time was observed beyond 512 environments, where minimum convergence time is observed, indicating that this scale represents a practical optimum for SHAC. Finally, DroQ benefits from parallelization less directly than on-policy algorithms, as its off-policy design emphasizes efficient use of replayed experience. No convergence was observed within 100 episodes with environment counts fewer than 512, whereas larger counts enabled convergence. Wall-clock training time for 100 episodes stayed near 12 minutes, with convergence improving to 8 minutes at 4096 environments.

\subsubsection{Effect of Attitude Representation}

\begin{figure}[t]
    \centering
    \includegraphics[width=\columnwidth]{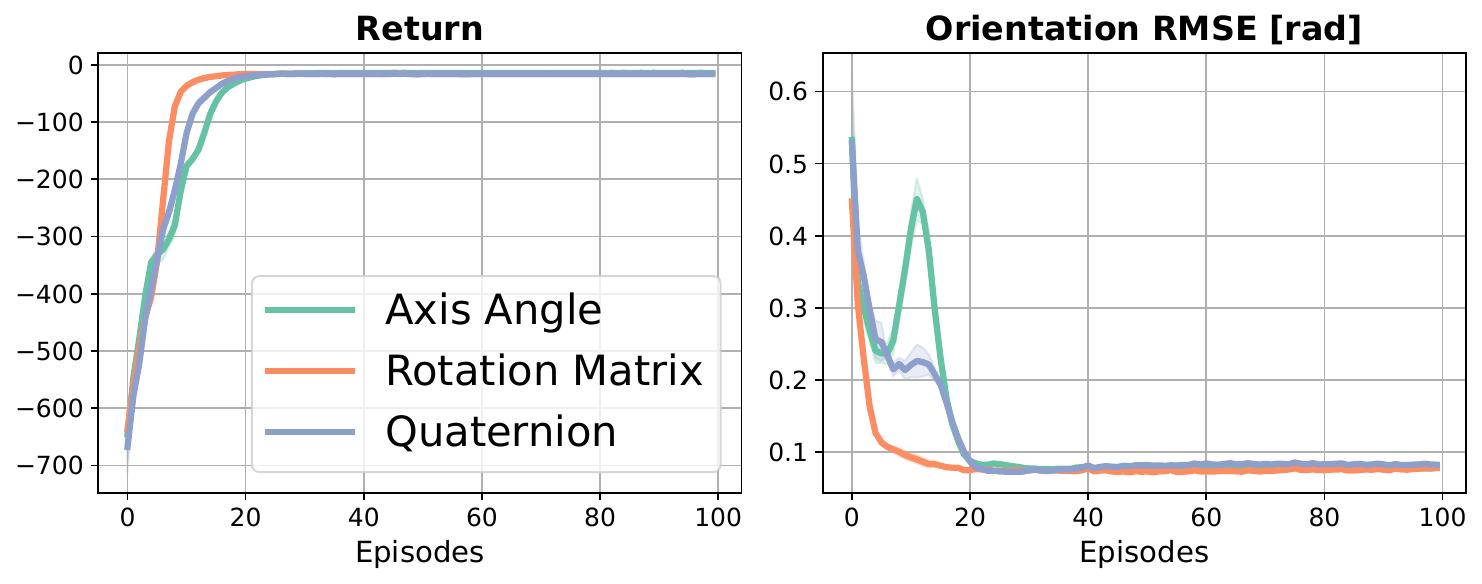}
    \caption{Effect of attitude representation on training performance and attitude control performance (trained with PPO).}
    \label{fig:att_rep}
\end{figure}

\begin{figure*}[htpb]
    \centering
    \includegraphics[width=\textwidth]{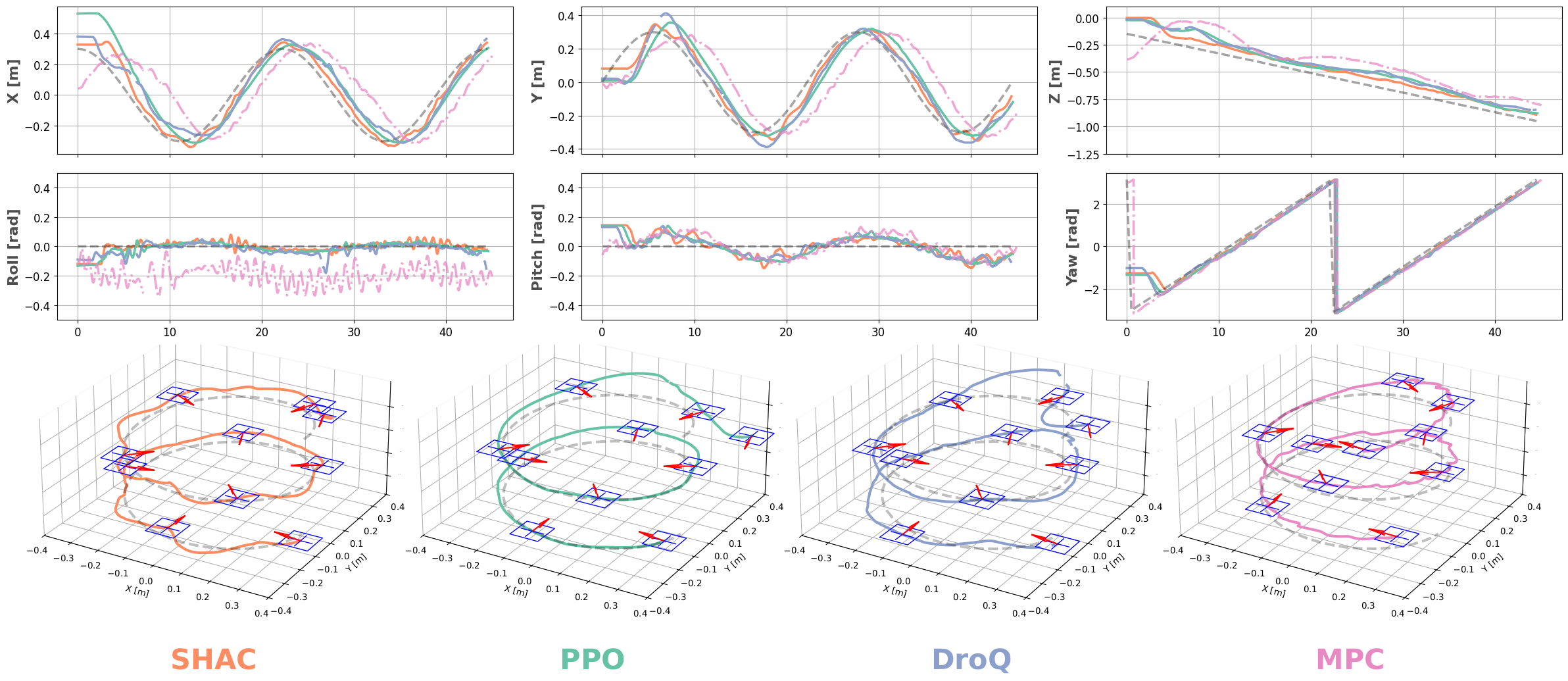}
    \caption{Helix trajectory experiment results. Per-axes graphs (top), 3D trajectories (bottom), red arrows represent vehicle heading.}
    \label{fig:control_helix}
\end{figure*}

An important design choice is how to represent the vehicle's orientation in the observation space. In principle, the choice of representation can influence both learning stability and control performance drastically, since different parameterizations of $SO(3)$ have different numerical properties. As shown in \cite{Zhou2018OnTC}, all representations under four or fewer dimensions are discontinuous, which will hinder learning, especially if one is interested in covering the full rotation space. This is where using errors, rather than the absolute attitude, becomes advantageous. In our formulation, orientation is expressed as a relative error with respect to the reference. This means the policy only needs to learn to reduce local deviations, which are numerically similar across representations in the moderate error regime. During training, we only sample initial references in that regime.

\begin{figure}[htpb]
    \centering
    \includegraphics[width=\columnwidth, trim={5cm 0cm 5cm 0cm}, clip]{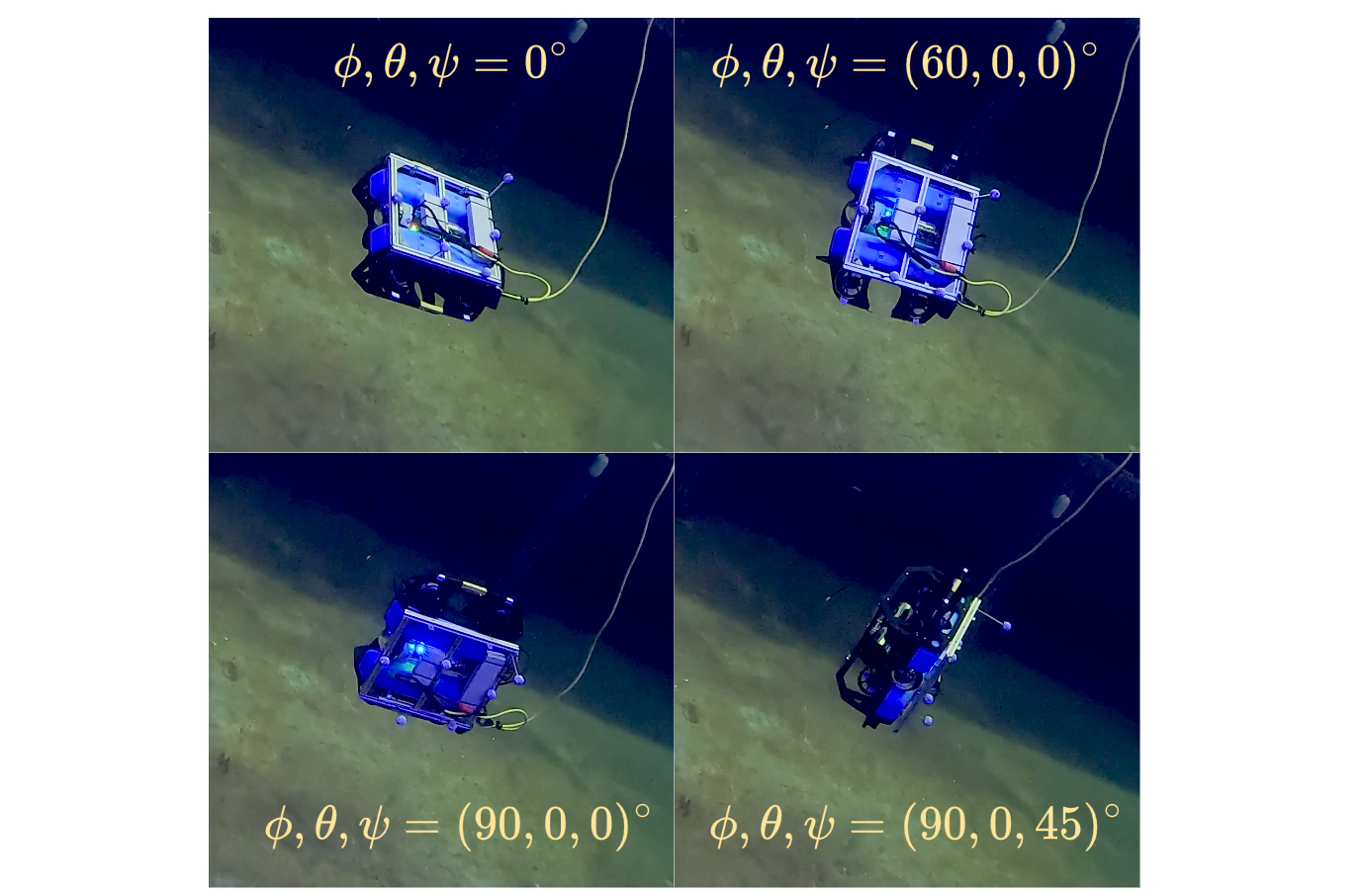}
    \caption{Attitude control demonstration. Vehicle holds large angle commands.}
    \label{fig:angle_exps}
\end{figure}

In order to determine whether the choice of attitude error representation has a meaningful effect on overall performance, we compared three common encodings: axis-angle, quaternion and rotation matrix. Our results are presented in Figure \ref{fig:att_rep}, which shows that all three representations achieve nearly identical asymptotic control performance. The only notable difference is a slight advantage in convergence speed when using the rotation matrix representation. Figure \ref{fig:angle_exps} provides an additional visual demonstration of the vehicle’s attitude control performance in real-world experiments.

\subsection{Real-World Experiments}

We evaluate the policies and the MPC baseline under realistic conditions by deploying them on a BlueROV2 Heavy platform. All experiments were conducted in the university’s indoor test tank facility. The vehicle’s motion was captured using a Qualisys underwater optical motion tracking system, which provided ground truth feedback used for both control and evaluation of tracking performance. Control integration was implemented using ROS 2. At runtime, the controllers output body-frame forces and torques, which are mapped to individual thruster commands through an allocation matrix, and then converted to PWM through an interpolation model.

\subsubsection{Center Locked Helix Tracking}

In this experiment, the vehicle was commanded to follow a helical trajectory while continuously orienting its body toward the centre of the helix. Trajectory waypoints are given sequentially every 0.25 seconds. The \emph{centre-locked} condition simultaneously excites the vehicle’s translational degrees of freedom and yaw, while also imposing indirect demands on pitch and roll through coupled dynamics. The results are presented in Figure \ref{fig:control_helix} and Table \ref{tab:helix_table}. All policies demonstrated strong control performance, validating that the trained controllers generalize effectively from simulation to real-world deployment. Among them, SHAC achieves the lowest 3D position RMSE of $0.099$ m and an attitude RMSE of $9.79^\circ$. PPO and DroQ yield comparable performance marginally worse than of SHAC's, while MPC shows the largest errors in both position and attitude. 

\begin{table}[t]
\centering
\caption{RMSEs for the centre locked helix experiment.}
\label{tab:helix_table}
\renewcommand{\arraystretch}{1.0}
\setlength{\tabcolsep}{3pt}
\small
\begin{tabular}{lcccccccc}
\toprule
\textbf{Method} & 
$x$ & $y$ & $z$ & 3D & 
Roll & Pitch & Yaw & Att. \\
 & [m] & [m] & [m] & [m] & [rad] & [rad] & [rad] & [deg] \\
\midrule
SHAC & \textbf{0.049} & \textbf{0.052} & \textbf{0.070} & \textbf{0.099} & \textbf{0.025} & 0.064 & \textbf{0.253} & \textbf{9.79} \\
PPO  & 0.093 & 0.090 & 0.100 & 0.164 & 0.024 & 0.071 & 0.279 & 12.05 \\
DroQ & 0.088 & 0.074 & 0.107 & 0.157 & 0.036 & \textbf{0.064} & 0.272 & 11.16 \\
MPC  & 0.200 & 0.172 & 0.173 & 0.315 & 0.207 & 0.076 & 0.291 & 17.51 \\
\bottomrule
\end{tabular}
\end{table}

\subsubsection{Disturbance Rejection}
To assess robustness against external perturbations, we performed disturbance–rejection trials in which the vehicle was commanded to maintain a fixed 6-DOF pose, and impulsive disturbances were manually applied by pulling the tether in different directions at varying times.

\begin{figure*}[t]
    \centering
    \includegraphics[width=\textwidth]{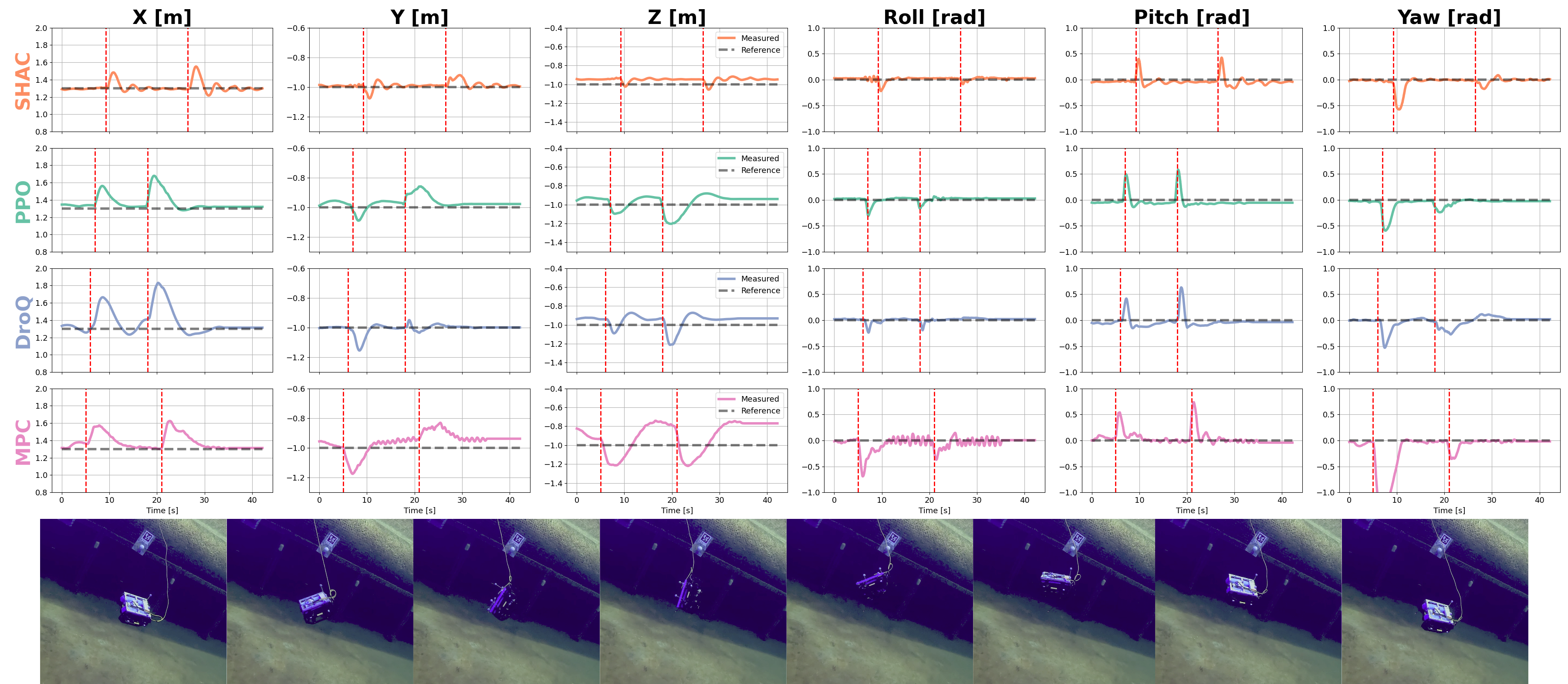}
    \caption{Disturbance rejection responses. Red dashed lines indicate disturbance instances. Images at the bottom row presents one such instance.}
    \label{fig:dist_control}
\end{figure*}

Table \ref{tab:dist_table} and Figure \ref{fig:dist_control} present experiment results. All three policies successfully attenuated the disturbances and restored the vehicle to the commanded setpoint. Overall, the SHAC policy achieved the lowest tracking error across all metrics. Its 3D position RMSE remained within $0.084,\text{m}$, compared to $0.150,\text{m}$ for PPO, $0.192,\text{m}$ for DroQ, and $0.232,\text{m}$ for the MPC baseline. The advantage is particularly pronounced in the vertical axis ($z$), where SHAC reduced the error by more than half compared to MPC. In terms of attitude, SHAC also demonstrated superior disturbance rejection, with an RMSE of only $8.54^{\circ}$. PPO and DroQ exhibited moderate degradation ($10.0^{\circ}$ and $9.7^{\circ}$, respectively), while MPC showed the weakest robustness with an error of $22.2^{\circ}$. 

\begin{table}[htbp]
\centering
\caption{RMSEs for disturbance rejection experiments.}
\renewcommand{\arraystretch}{1.0}
\label{tab:dist_table}
\setlength{\tabcolsep}{3pt}
\small
\begin{tabular}{lcccccccc}
\toprule
\textbf{Method} & 
$x$ & $y$ & $z$ & 3D & 
Roll & Pitch & Yaw & Att. \\
 & [m] & [m] & [m] & [m] & [rad] & [rad] & [rad] & [deg] \\
\midrule
SHAC & \textbf{0.059} & \textbf{0.025} & \textbf{0.054} & \textbf{0.084} & \textbf{0.041} & \textbf{0.091} & \textbf{0.114} & \textbf{8.54} \\
PPO  & 0.112 & 0.050 & 0.086 & 0.150 & 0.052 & 0.114 & 0.128 & 10.00 \\
DroQ & 0.170 & 0.031 & 0.083 & 0.192 & \textbf{0.041} & 0.118 & 0.117 & 9.70 \\
MPC  & 0.118 & 0.078 & 0.184 & 0.232 & 0.140 & 0.141 & 0.352 & 22.20 \\
\bottomrule
\end{tabular}
\end{table}

\section{Discussion}

\subsection{Limitations}

\begin{figure}[htpb]
    \centering
    \includegraphics[width=\columnwidth]{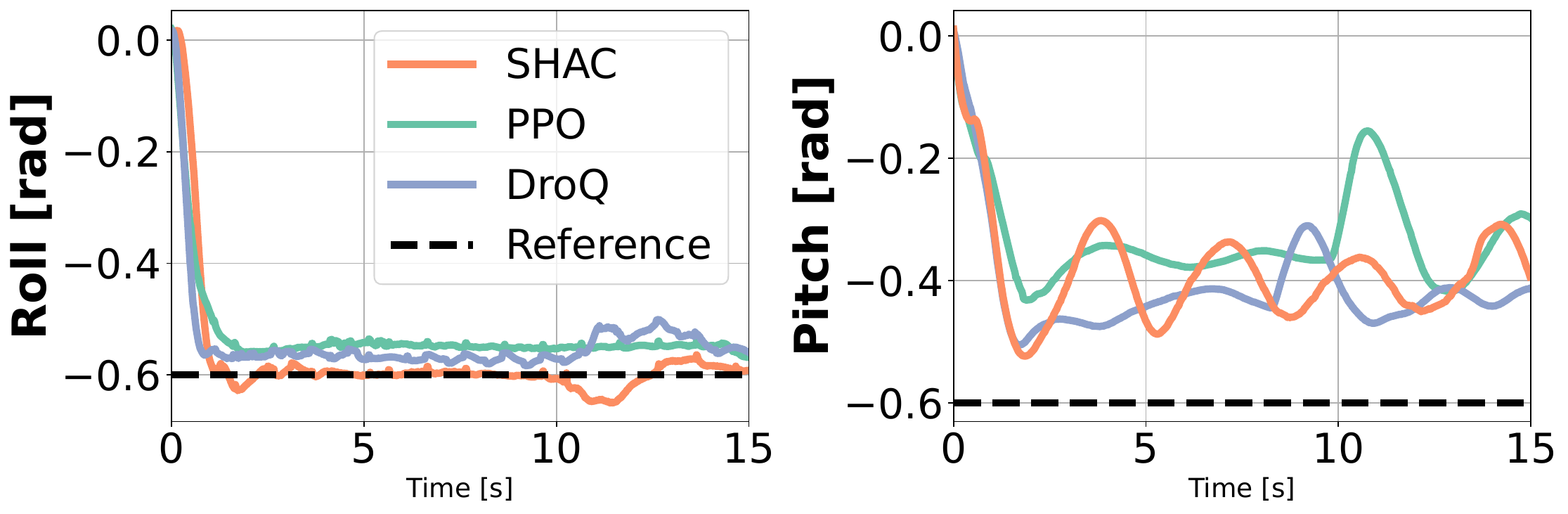}
    \caption{Step response for roll and pitch.}
    \label{fig:control_step_angle}
\end{figure}

While all controllers demonstrated strong performance across our evaluations, limitations were observed along the pitch axis with real world experiments. Specifically, when large pitch angles were commanded, we consistently measured a small but persistent steady-state error. This is illustrated in Figure \ref{fig:control_step_angle}, reporting step responses for roll and pitch axes with a step magnitude of $0.6\text{rad}$. Roll control converged without residual error, whereas pitch control exhibited an offset. We attribute this to a sim-to-real gap: in simulation, hydrodynamic models assumed perfect alignment of the centers of gravity and buoyancy, whereas in the real vehicle even small misalignments introduce a restoring moment about the pitch axis. Such unmodeled static moments likely explain the residual error. As future work, we plan to extend the simulator with offset buoyancy effects and thereby validate this hypothesis.

\subsection{Caveat on Reward Shaping and Controller Tuning}
It is important to emphasize that the results presented in this work do not represent the absolute performance limits of the evaluated controllers. In reinforcement learning, different algorithmic paradigms exhibit distinct sensitivities to the reward structure, meaning that control performance can be further improved by carefully engineering task-specific reward terms. Likewise, the MPC baseline can be improved through better model identification, or more sophisticated optimization formulations.

However, our goal was not to hand-engineer the best possible reward functions or controller settings for each algorithm. Instead, the reward function was engineered to enable all reinforcement learning algorithms to reach a competitive level of performance, rather than to maximize the absolute performance of a single method. In practice, this yielded broadly comparable levels of performance across the evaluated controllers, with SHAC exhibiting slightly lower errors than the other methods. 

\section{Conclusion}
This work introduced a GPU-accelerated reinforcement learning framework for 6-DOF position control of underwater vehicles, combining differentiable simulation in MJX with large-scale parallelization and jit compilation. By jointly compiling physics and learning updates, we reduced training times up to a couple of minutes while obtaining policies that reliably achieve accurate 6-DOF control. All RL policies demonstrated precise trajectory tracking and disturbance rejection. These results establish that fast policy learning and robust sim-to-real transfer are feasible for full 6-DOF underwater vehicle control. More broadly, the GPU-accelerated MJX–JAX pipeline demonstrates a general approach for fast policy learning with differentiable physics, with potential impact beyond underwater robotics.

\bibliographystyle{IEEEtran}
\bibliography{IEEEabrv,bib}

\end{document}